\documentclass[11pt,twocolumn,twoside]{IEEEtran}
\usepackage{amsmath}
\usepackage[margin=0.75in,headheight=0.45in]{geometry}
\usepackage[pdftex]{epsfig}
\usepackage{amsfonts}
\usepackage{amssymb}
\usepackage{fancyhdr}

\usepackage{cite}
\usepackage{amsmath,amssymb,amsfonts}
\usepackage{graphicx}
\usepackage{textcomp}
\usepackage{xcolor}
\usepackage{booktabs} 
\usepackage[linesnumbered,ruled,vlined]{algorithm2e}
\usepackage{booktabs}
\usepackage{amsmath}
\usepackage{graphicx}
\usepackage{subfigure}
\usepackage{cite}
\usepackage{multirow}
\usepackage{amsfonts,bm}
\usepackage{color}
\usepackage{braket}
\usepackage{mathtools}
\usepackage{enumerate}
\usepackage{enumitem}
\usepackage{empheq}
\usepackage{calc}
\usepackage{balance}
\usepackage{dsfont}
\usepackage{calc}

\include{graphicsx}

\begin{document}
\title{\vspace{0.2in}\sc Physics Guided Recurrent Neural Networks for modeling dynamical systems: application to monitoring water temperature and quality in lakes }
\author{Xiaowei Jia$^1$\thanks{$^1$ University of Minnesota, \{jiaxx221,karpa009,willa099, stei0062,kumar001\}@umn.edu. $^2$United States Geologic Survey, jread@usgs.gov. $^3$University of Wisconsin-Madison, Center for Limnology, \{pchanson,hdugan\}@wisc.edu.}, Anuj Karpatne$^1$, Jared Willard$^1$, Michael Steinbach$^1$, Jordan Read$^2$, Paul C Hanson$^3$, Hilary A Dugan$^3$, Vipin Kumar$^1$}

\maketitle
\thispagestyle{fancy}
\begin{abstract}
In this paper, we introduce a novel framework for combining
scientific knowledge within physics-based models and recurrent neural networks to advance scientific discovery in many dynamical systems. We will first describe the use of outputs from physics-based models in learning a hybrid-physics-data model. Then, we further incorporate physical knowledge in real-world dynamical systems as additional constraints for training recurrent neural networks. We will apply this approach on modeling lake temperature and quality where we take into account the physical constraints along both the depth dimension and time dimension.  
By using scientific knowledge to guide the construction and learning the data-driven model, we demonstrate that this method can achieve better prediction accuracy as well as scientific consistency
of results.
\end{abstract}

\section{Introduction}
Physics-based  models  of  dynamical  systems  are  often  used  to  study  engineering  and environmental  systems.  Despite  their  extensive  use,  these  models  have  several  well-known  limitations  due  to incomplete or  inaccurate representations of  the physical  processes  being modeled. Given rapid data growth due to  advances  in  sensor  technologies,  there  is  a  tremendous  opportunity  to  systematically  advance  modeling  in these  domains  by  using  machine  learning  (ML)  methods.  However,  direct  application of  black-box  ML  models  to  a  scientific  problem  encounters  several  major  challenges.  First,  in  the  absence  of adequate  information  about  the  physical  mechanisms  of  real-world  processes,  ML  approaches  are  prone  to  false discoveries  and  can  also  exhibit  serious  inconsistencies  with  known  physics.  This  is  because  scientific  problems often  involve  complex  spaces  of  hypotheses  with  non-stationary  relationships  among  the  variables  that  are difficult  to  capture  solely  from  the  data.  Second,  black-box  ML  models  suffer  from  poor  interpretability  since they are  not  explicitly  designed  for  representing  physical  relationships  and  providing  mechanistic  insights.  Third, the  data  available  for  several  scientific  problems  are  far  smaller  than  what  than  what  is  needed  to  effectively train  advanced  ML  models.  Leveraging  physics  will  be  key  to  constrain  hypothesis  spaces  to  do  ML  in  such small  sample  regimes.  Hence,  neither  an  ML-only  nor  a  physics-only  approach  can  be  considered  sufficient  for knowledge  discovery  in  complex  scientific  and  engineering  applications.  Instead,  there  is  a  need  to  explore  the continuum  between  physics-based  and  ML  models, where  both  physics  and  data  are  integrated  in  a synergistic  manner.  Next we outline issues involved in building such a hybrid model that is already beginning to show great promise~\cite{karpatne2017physics}. 

In science and engineering applications, a physical model often predicts values of many different variables.  Machine learning models can also generate predictions for many different variables (e.g., by having multiple nodes in the output layer of a neural network).  Most ML algorithms make use of a loss function that captures the difference between predicted and actual (i.e., observed values) to guide the search for parameter values that attempts to minimize this loss function.  Although, such empirical models are often used in many scientific communities as alternatives to physical models, they fail to take in to account many physical aspects of modeling.  In the following we list some of these.

In science and engineering applications, all errors (i.e., difference between predicted and observed values) may not be equally important.  For example, for the lake lake temperature monitoring application, accuracy at surface and at high depth can be more important than error at the middle levels of the lake. 

Instead of 
minimizing the difference between predicted and observed values, it may be more important to optimize the prediction of a different physical quantity, which can be computed from the observed or predicted values.  For example, for certain lake temperature monitoring applications, the ability to correctly predict the depth of thermocline (i.e., depth at which temperature gradient is maximum) can be more important than correctly predicting the temperature profile at all depths. 

Values of different variables predicted by a science and engineering model may have certain relationships (guided by physical laws) across space and time.  For example, in the lake temperature monitoring application, predicted values of the temperature at different depths should be such that denser water is at lower depth (note that water is heaviest at 4 degree centigrade).  As another example, changes in temperature profile across time involves transfer of energy and mass across different layers of a lake that must be conserved according to physical laws. 

In this paper, we propose a novel framework, Physics-Guided Recurrent Neural Networks (PGRNN) that can incorporate many of these physical aspects by designing non-standard loss functions and new architectures.  We motivate and illustrate these ideas in the context of monitoring temperature and water quality in lakes, but they are applicable to a broad range of science and engineering problems.

\section{Physics-Guided Recurrent Neural Networks }
\subsection{Long-Short Term Memory}
We first briefly describe the structure of the Long-Short Term Memory (LSTM) model. Given the input  $x^t$ at every time step, the LSTM model generates hidden representation/embeddings $h^t$ at every time step, which are then used for prediction. In essense, the LSTM model defines a transition relationship for hidden representation $h^{{t}}$ 
through an LSTM cell, 
which takes the input of features $x^{t}$ at the current time step and also the inherited information from previous time steps.

Each LSTM cell contains a cell state $c^t$, which serves as a memory and allows the hidden units $h^t$ to reserve information from the past. The cell state $c^t$ is generated by combining $c^{t-1}$, $h^{t-1}$, and the input features at $t$. Hence, the transition of cell state over time forms a memory flow, which enables the modeling of long-term dependencies. Specifically, we first generate a new candidate cell state $\tilde{c}^t$ by combining $x^t$ and $h^{t-1}$ into a $\text{tanh}(\cdot)$ function, as follows:
\begin{equation}
\footnotesize
\begin{aligned}
\tilde{c}^t &= \text{tanh}(W^c_h h^{t-1} + W^c_x x^t),
\end{aligned}
\end{equation}
where $W^c_h\in \mathds{R}^{H\times H}$ and $W^c_x\in \mathds{R}^{H\times D}$ denote the weight parameters used to generate candidate cell state. Hereinafter we omit the bias terms as they can be absorbed into weight matrices. 
Then we generate a forget gate layer $f^t$, an input gate layer $g^t$, and an output gate layer, as:
\begin{equation}
\footnotesize
\begin{aligned}
f^t &= \sigma(W^f_h h^{t-1} + W^f_x x^t),\\
g^t &= \sigma(W^g_h h^{t-1} + W^g_x x^t),\\
o^t &= \sigma(W^o_h h^{t-1} + W^o_x x^t).
\end{aligned}
\end{equation}

Then we compute the new cell state and the hidden representation as: 
\begin{equation}
\begin{aligned}
\small
c^t &= f^t\otimes c^{t-1}+g^t\otimes\tilde{c}^t,\\
h^t &= o^t\otimes \text{tanh}(c^t).
\end{aligned}
\end{equation}


\subsection{Hybrid-physics-data Model}
Then we construct a hybrid model in two steps. First, we propose to integrate the predicted outputs $Y_{phy}$ from physics-based models as input to the LSTM model. If the goal of the traditional data science model is to learn a mapping $f_{RNN}: D\rightarrow Y$, the hybrid model can be represented as $f_{HPD}: [D,Y_{phy}]\rightarrow Y$, where $Y_{phy}$ is the output from the physics-based model.  In physics-based models, the use of $Y_{phy}$ by itself may provide an incomplete representation of the target variable due to simplified or missing physics. By including $Y_{phy}$ as part of the input for the data science model, we aim to fill in the gap between $Y_{phy}$ and true observations while maintaining the physical knowledge in  $Y_{phy}$. 

Second, we use $Y_{phy}$ to refine the training loss for the time steps with missing observations. The effective learning of LSTM requires frequently collected data. However, real-world observations can be missing or noisy on certain dates. Therefore, the use of $Y_{phy}$ on those missing dates can provide a complete temporal trajectory for training LSTM.   

\begin{figure} [!h]
\centering
\includegraphics[width=0.8\columnwidth]{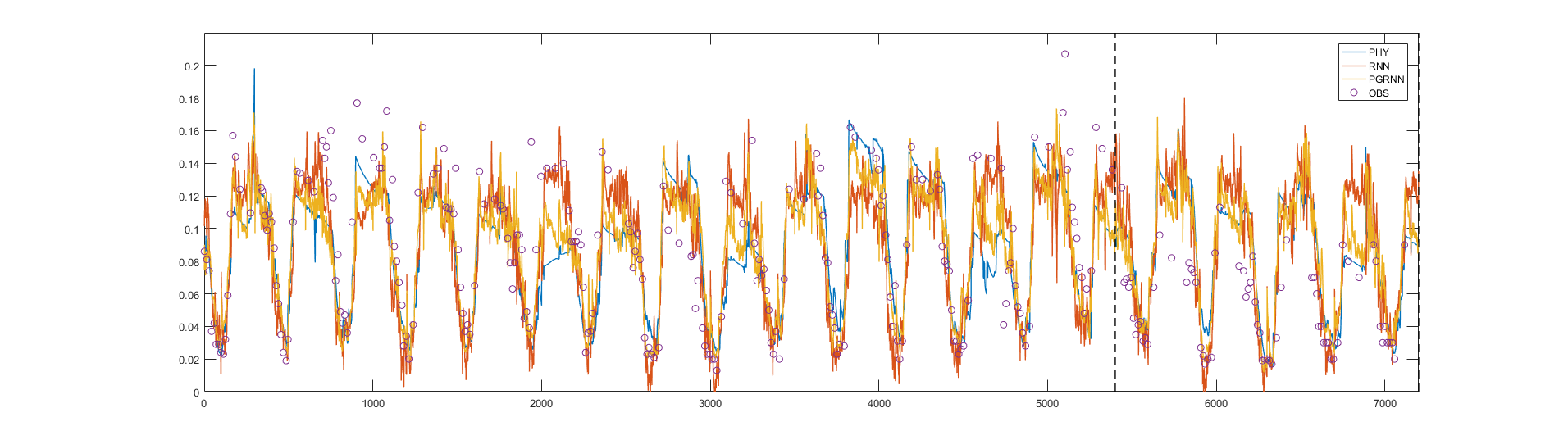}
\caption{The predicted surface phosphorus concentration for Lake Mendota from 1996 to 2015. The part before the dashed line is used for training and the part after the dashed line is used for testing. }
\label{pho}
\end{figure}

We apply this hybrid model in predicting phosphorus concentration over time for Lake Mendota, as shown in Fig.~\ref{pho}. In Table~\ref{pho_val}, we can see that PGRNN can significantly improve the prediction accuracy. 

\begin{table}[!h]
\footnotesize
\newcommand{\tabincell}[2]{\begin{tabular}{@{}#1@{}}#2\end{tabular}}
\centering
\caption{The performance of each method (in terms of RMSE for the entire test period, winter, and summer) in predicting surface phosphorus concentration for Lake Mendota.  }
\begin{tabular}{lccc}
\hline
Method& Overall & Winter & Summer \\ \hline 
PHY & 0.0266 &  0.0306 & 0.0237\\  
RNN & 0.0247 & 0.0312 & 0.0190\\  
PGRNN & 0.0237 & 0.0297 & 0.0209\\ \hline
\end{tabular}
\label{pho_val}
\end{table}

\subsection{Physical Constraints}

Having described the hybrid model, we now add additional constraints for training this model so that the predictions are physically consistent. To better illustrate this, we consider the example of lake temperature monitoring. We introduce two constraints along the depth dimension and the time dimension, respectively. 

\textbf{Density-depth relationship: }
It is known that the density of water monotonically increases with depth. Also, the temperature, $Y$ , and density, $\rho$, of water are related to each other according to the following known physical equation~\cite{martin2018hydrodynamics}:
\begin{equation}
\footnotesize
    \rho = 1000\times (1-\frac{(Y+288.9414)\times (Y-3.9863)^2}{508929.2\times (Y+68.12963)}).
\label{td}
\end{equation}

We first transform the values of predicted temperature into the density values according to Eq.~\ref{td}. Then, we add an extra penalty for violation of density-depth relationship. 

In Table~\ref{perf_mendota}, we report some preliminary results. Our dataset is comprised of 13,543  observations from 30 April 1980 to  02 Nov 2015. We use 2/3 of data of training while testing on the remaining 1/3 data. For each observation, we used a set
of 11 meteorological drivers as input variables, including wind speed, rain, freezing conditions, long-wave and short-wave radiation, etc. 

\begin{figure} [!h]
\centering
\subfigure[]{ \label{fig:a}{}
\includegraphics[width=0.45\columnwidth]{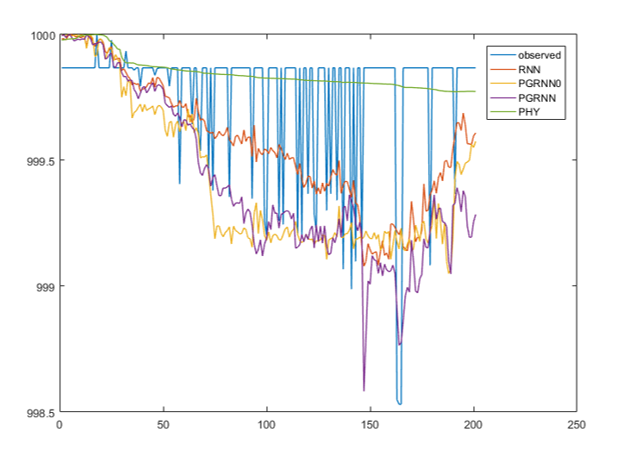}
}\hspace{-.07in}
\subfigure[]{ \label{fig:b}{}
\includegraphics[width=0.45\columnwidth]{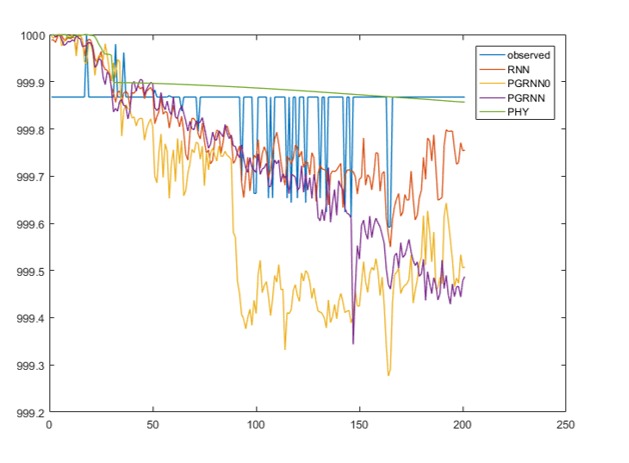}
}\hspace{-.07in}
\caption{The predicted density value from 05-Apr-1993 to 22-Oct-1993 for Lake Mendota at depth (a) 14 m and (b) 25 m.}
\label{temporal_val}
\end{figure}

It can be seen that PGRNN (with density-depth constraint) outperforms PGRNN0 (without density-depth constraint) for both RMSE and Phy-inconsistency. Also, the comparison between RNN (LSTM networks) and ANN shows that the modeling of temporal transition can help better capture the temperature change over time. From a temporal perspective, we observe from Fig.~\ref{temporal_val} that PGRNN can better capture the changes at certain depths where physics-based model and traditional RNN cannot achieve reasonable accuracy. 

\begin{table}[!h]
\footnotesize
\newcommand{\tabincell}[2]{\begin{tabular}{@{}#1@{}}#2\end{tabular}}
\centering
\caption{The performance of each method in predicting water temperature for Lake Mendota. The phy-inconsistency represents the average number of violations for density-depth relationship at every date and every depth. PHY represents physics-based GLM model. }
\begin{tabular}{lcc}
\hline
Method& RMSE & Phy-inconsistency \\ \hline 
PHY & 2.6544 &  0.0051\\  
ANN & 1.8830  &  0.1920\\  
RNN & 1.6042 & 0.2024\\  
PGRNN0 & 1.6068 & 0.1798\\ 
PGRNN & 1.4791 & 0.0732\\ \hline
\end{tabular}
\label{perf_mendota}
\end{table}

\textbf{Energy conservation: }

The temperature change in lake water is caused by the energy flow over time. The lake energy budget is a balance between incoming energy fluxes and heat losses from the lake. A mismatch in losses and gains results in a temperature change - more gains than losses will warm the lake, and more losses than gains will cool the lake. 

Given the temporal modeling structure in the LSTM model, we add constraint on the predicted temperature over time such that the change of volume-average temperature is consistent to the energy gain/loss. 

\begin{figure} [!h]
\centering
\subfigure[]{ \label{fig:a}{}
\includegraphics[width=0.4\columnwidth]{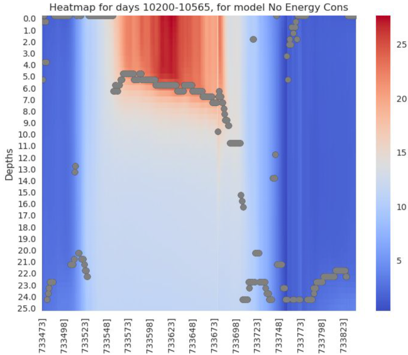}
}\hspace{-.07in}
\subfigure[]{ \label{fig:b}{}
\includegraphics[width=0.4\columnwidth]{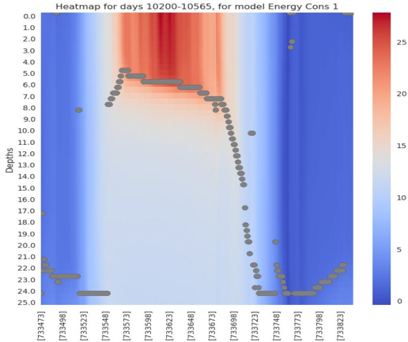}
}\hspace{-.07in}
\caption{The thermocline (grey dots) detected from predicted temperatures using (a) PGRNN without energy conservation constraint, and (b) PGRNN with energy conservation constraint. The x-axis represents the date from March 5th 2008 to March 4th 2009.}
\label{EC}
\end{figure}

In Fig.~\ref{EC}, we show the thermocline depth detected in Lake Mendota. It can be seen that  the detected thermocline position evolves more smoothly over time after we integrate the energy conservation constraint.

\bibliographystyle{ieeetr}
\bibliography{ci_references}

\end{document}